\documentclass[10pt,final,journal]{IEEEtran} 
\newcommand{\xz}{\textcolor{red}}
\newcommand{\xzb}{\textcolor{blue}}
\newcommand{\xzg}{\textcolor{green}}

\usepackage{lineno}
\usepackage[colorlinks,
linkcolor=red,       
anchorcolor=green,  
citecolor=green,        
]{hyperref}
\usepackage{moreverb}
\usepackage{fancyhdr}
\usepackage{calc}
\usepackage{float}
\usepackage{subfigure}
\usepackage{color}
\usepackage[table]{xcolor}
\usepackage{amsmath}
\usepackage{arydshln}
\usepackage{ulem} 

\usepackage{graphicx}
\usepackage{epstopdf}

\usepackage{bm}
\usepackage{setspace}
\newcommand{\tabincell}[2]{\begin{tabular}{@{}#1@{}}#2\end{tabular}}

\usepackage{enumerate}
\usepackage{diagbox}

\modulolinenumbers[5]

\usepackage{array}
\newcommand{\PreserveBackslash}[1]{\let\temp=\\#1\let\\=\temp}
\newcolumntype{C}[1]{>{\PreserveBackslash\centering}p{#1}}
\newcolumntype{R}[1]{>{\PreserveBackslash\raggedleft}p{#1}}
\newcolumntype{L}[1]{>{\PreserveBackslash\raggedright}p{#1}}

\usepackage{multirow}
\usepackage[figuresright]{rotating}
\usepackage{cite}

\usepackage{amsthm,amsmath,amssymb}

\usepackage{mathrsfs}
\usepackage{lipsum} 

\usepackage{bbding}

\usepackage{makecell}
\usepackage[numbers,sort&compress]{natbib}

\usepackage[table]{xcolor}

\ifCLASSINFOpdf

\else

\fi

\begin{document}

\title{Self-Supervised RGB-T Tracking with Cross-Input Consistency}

	\author{Xingchen~Zhang$^*$, \IEEEmembership{\textit{Member}, IEEE,} Yiannis Demiris, \IEEEmembership{Senior Member, IEEE}
			\thanks{X.~Zhang and Y.~Demiris are
				with the Personal Robotics Laboratory, Department of Electrical and Electronic Engineering,
				Imperial College London, London SW7 2AZ, U.K. (e-mail: xingchen.zhang@imperial.ac.uk, y.demiris@imperial.ac.uk)
			 \newline	$^*$ Corresponding author: Xingchen Zhang 	}
	}

\markboth{Journal of \LaTeX\ Class Files,~Vol.~XX, No.~XX, 2023}%
{Shell \MakeLowercase{\textit{et al.}}: Bare Demo of IEEEtran.cls for Computer Society Journals}

\IEEEtitleabstractindextext{%
\begin{abstract}
    In this paper, we propose a self-supervised RGB-T tracking method.~Different from existing deep RGB-T trackers that are using a large number of annotated RGB-T image pairs for training, our RGB-T tracker is trained using unlabeled RGB-T video pairs in a self-supervised manner.~We propose a novel cross-input consistency-based self-supervised training strategy based on the idea that tracking can be performed using different inputs.~Specifically, we construct two distinct inputs using unlabeled RGB-T video pairs.~We then track objects using these two inputs to generate results, based on which we construct our cross-input consistency loss.~Meanwhile, we propose a re-weighting strategy to make our loss function robust to low-quality training samples.~We build our tracker on a Siamese correlation filter network.~To the best of our knowledge, our tracker is the first self-supervised RGB-T tracker.~Extensive experiments on two public RGB-T tracking benchmarks demonstrate that the proposed training strategy is effective.~Remarkably, despite training only with a corpus of unlabeled RGB-T video pairs, our tracker outperforms seven supervised RGB-T trackers on the GTOT dataset.
\end{abstract}

}

\maketitle
\IEEEdisplaynontitleabstractindextext
\IEEEpeerreviewmaketitle


\section{Introduction}
\IEEEPARstart{O}{bject} tracking is an important task and has many applications in areas such as robots and surveillance.~In recent years, many tracking algorithms have been proposed, and tracking performance has witnessed a significant improvement.~However, most visual trackers operate on RGB images.~The performance of these trackers degrades significantly when RGB images are not reliable (e.g., under poor lighting conditions), limiting their practical applications.

\begin{figure}
	\centering
	\includegraphics[width=0.475\textwidth]{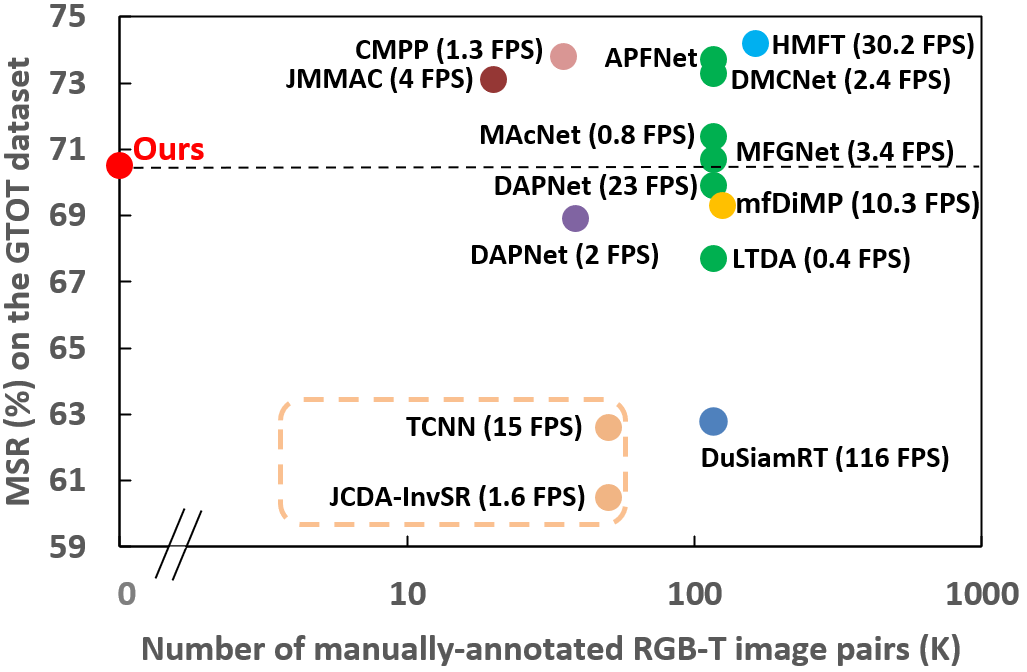}
	\caption{Performance v.s.~required labeled RGB-T image pairs in training.~Although our RGB-T tracker does not need any labeled training data, it outperforms seven supervised RGB-T trackers on the GTOT dataset \cite{li2016learning}.~Note that the exact number of labeled data for TCNN \cite{li2018fusing} and JCDA-InvSR  \cite{kang2019grayscale} (both are supervised trackers) is not mentioned in their papers.
	}
	\label{fig:msr-size}
\end{figure}

To improve tracking performance, researchers have used thermal images and RGB images together to perform RGB-T tracking \cite{li2019rgb,zhang2020object,zhang2019siamft,zhang2020dsiammft,zhang2020objectfusion,zhang2021jointly,lu2021rgbt}.~This is based on the fact that thermal images are insensitive to illumination changes while RGB images contain more texture details \cite{zhang2020vifb}.~Although many efforts have been put into developing deep learning-based RGB-T trackers and RGB-T tracking performance has been significantly improved, existing deep RGB-T trackers  \cite{zhang2022visible,wang2020cross,lu2022duality,zhang2021jointly} require a large number of annotated RGB-T image pairs, as shown in Fig.~\ref{fig:msr-size}.

It is well-known that annotation is time-consuming and expensive.~Some unsupervised single object trackers have been proposed to avoid the need for annotations.~For example, Wang et al.~\cite{wang2019unsupervised, wang2021unsupervised} and Zhu et al.~\cite{zhu2021contrastive}
use a cycle consistency based on forward-backward tracking to train trackers.~There are also some trackers using very spare annotation in training, e.g., annotation in the initial frame \cite{yuan2020self,yuan2021self,zheng2021learning}.~However, \textit{all unsupervised trackers use only single-modal images}, namely, either use RGB images \cite{vondrick2018tracking, wang2019unsupervised, wang2021unsupervised, yuan2020self, yuan2021self, li2021self, zheng2021learning,shen2022unsupervised, wu2021progressive} or thermal images \cite{huang2022thermal}.

In this paper, we propose a self-supervised RGB-T tracker that does not need any manual annotations in training.~To achieve this, we propose a \textit{cross-input consistency}-based training strategy to exploit temporal information in unlabeled RGB-T videos.~Our intuition resides on the observation that object tracking can be performed using different inputs.~As shown in Fig.~\ref{fig:idea}, given a target at frame $t$, we can track it to obtain its position at frame $(t+1)$ using different inputs (e.g., RGB images, thermal images, or a combination of them).~Ideally, if all tracking are successful, the tracking results in frame $(t+1)$ should be consistent.

We integrate our self-supervised training strategy into a Siamese-based discriminative correlation filter (DCF) framework.~In implementation, we construct two distinct inputs for tracking to build cross-input consistency.~This cross-input consistency, which is based on \textit{temporal information in unlabeled RGB-T video pairs}, can be used to guide the training of our RGB-T tracker.~In addition, because we do not want to use any manual annotations, we randomly initialize a bounding box in our training data.~Therefore, the training samples are usually noisy or have bad quality.~We propose a re-weighting strategy to re-weight our loss function to make our training easier and more effective.~In summary, the main contributions of this paper include:

\begin{figure}
	\centering
    \includegraphics[width=0.45\textwidth]{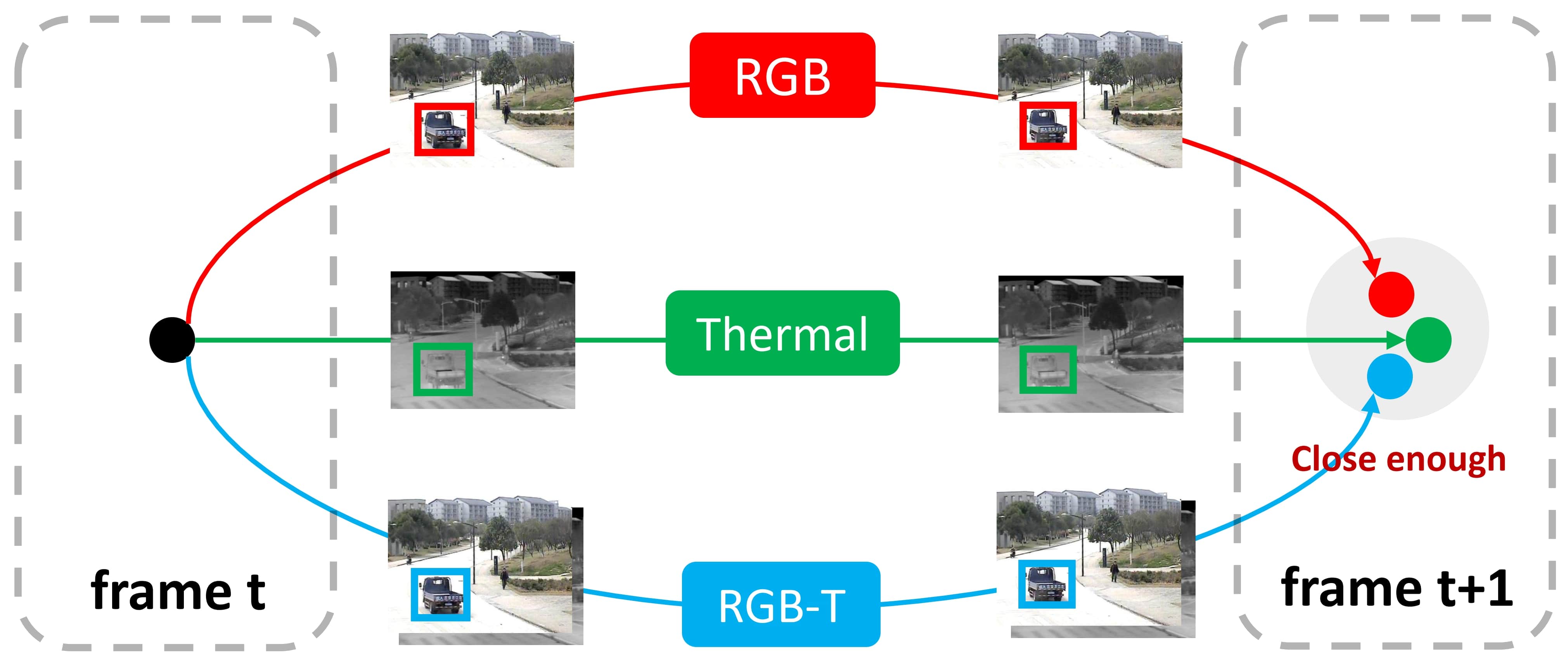}
	\caption{Our cross-input consistency is based on the observation that tracking can be performed using different inputs.~Ideally, the tracking results in frame $(t+1)$ obtained from different inputs should be close enough.}
	\label{fig:idea}
\end{figure}

\begin{itemize}
	\item We propose a self-supervised RGB-T tracker trained using RGB-T video pairs without  human annotations.
	
	\item We propose a cross-input consistency-based strategy to achieve self-supervised training.~We use 
    RGB images and thermal images to construct different inputs for tracking, based on which a cross-input consistency loss is constructed to guide training.

	\item We propose a re-weighting scheme to re-weight our loss function to make the training more effective.
	
	\item Extensive experiments on two RGB-T tracking benchmarks demonstrate the favorable performance of the proposed method and the potential of self-supervised RGB-T tracking.
\end{itemize}

The rest of this paper is organized as follows.~Section \ref{sec:related} introduces related work.~Then, Section \ref{sec:method} introduces the proposed method in detail, followed by the introduction to training data processing in Section  \ref{sec:training}.~Then, Section \ref{sec:experiments} presents results and Section \ref{sec:discussion} gives discussions.~Finally, Section \ref{sec:conclusion} concludes this paper.

\section{Related Work}
\label{sec:related}
\subsection{Single object tracking}

Single object tracking methods mainly include deep learning-based methods \cite{bhat2019learning,li2019target,wang2021unsupervised} and discriminative correlation filter (DCF)-based methods \cite{li2020autotrack,zheng2021mutation}.~Most trackers 
use RGB images as input and have a high requirement for good lighting conditions.~To make trackers insensitive to light conditions, some researchers performed tracking using thermal images \cite{felsberg2015thermal,liu2020learning}.~However, thermal images do not have enough texture details, leading to worse performance than RGB-based trackers when lighting conditions are good.

\subsection{RGB-T tracking}
To alleviate the issue of RGB-based and thermal-based trackers, researchers performed RGB-T tracking \cite{zhang2019object, zhang2020object, li2019rgb, zhang2022visible}.~For example, Zhang et al.~\cite{zhang2019object} proposed a pixel-level fusion-based RGB-T tracker.~In contrast, some RGB-T trackers are based on feature-level fusion \cite{li2020challenge, zhang2021learning, lu2021rgbt} or decision-level \cite{zhang2019decision} or combine several fusion levels \cite{zhang2022visible}.~The performance of RGB-T trackers have been significantly improved.~However, existing deep RGB-T trackers need a large number of RGB-T image pairs for training.

\subsection{Unsupervised object tracking}
Researchers have proposed unsupervised trackers to alleviate the need for annotations.~For example, Vondrick et al.~\cite{vondrick2018tracking} proposed to train an RGB tracker by colorizing videos.~Wang et al.~\cite{wang2019unsupervised, wang2021unsupervised} and Shen et al.~\cite{shen2022unsupervised} proposed to use cycle consistency to train an RGB tracker.~Yuan et al.~\cite{yuan2020self} and Shen et al.~\cite{shen2022unsupervised} further used region proposal network in the cycle consistency framework.~Some other unsupervised RGB trackers have also been proposed based on different ideas, such as cycle memory learning \cite{zheng2021learning}, crop-transform-paste operation \cite{li2021self}, and training using images and their cropped regions \cite{sio2020s2siamfc}.~In addition, unsupervised thermal tracker based on cycle consistency has also been proposed \cite{huang2022thermal}.~However, \textit{existing unsupervised trackers are limited to one single modality}, i.e., based on only RGB images or only thermal images.

\begin{figure*}
	\centering
	\includegraphics[width=\textwidth]{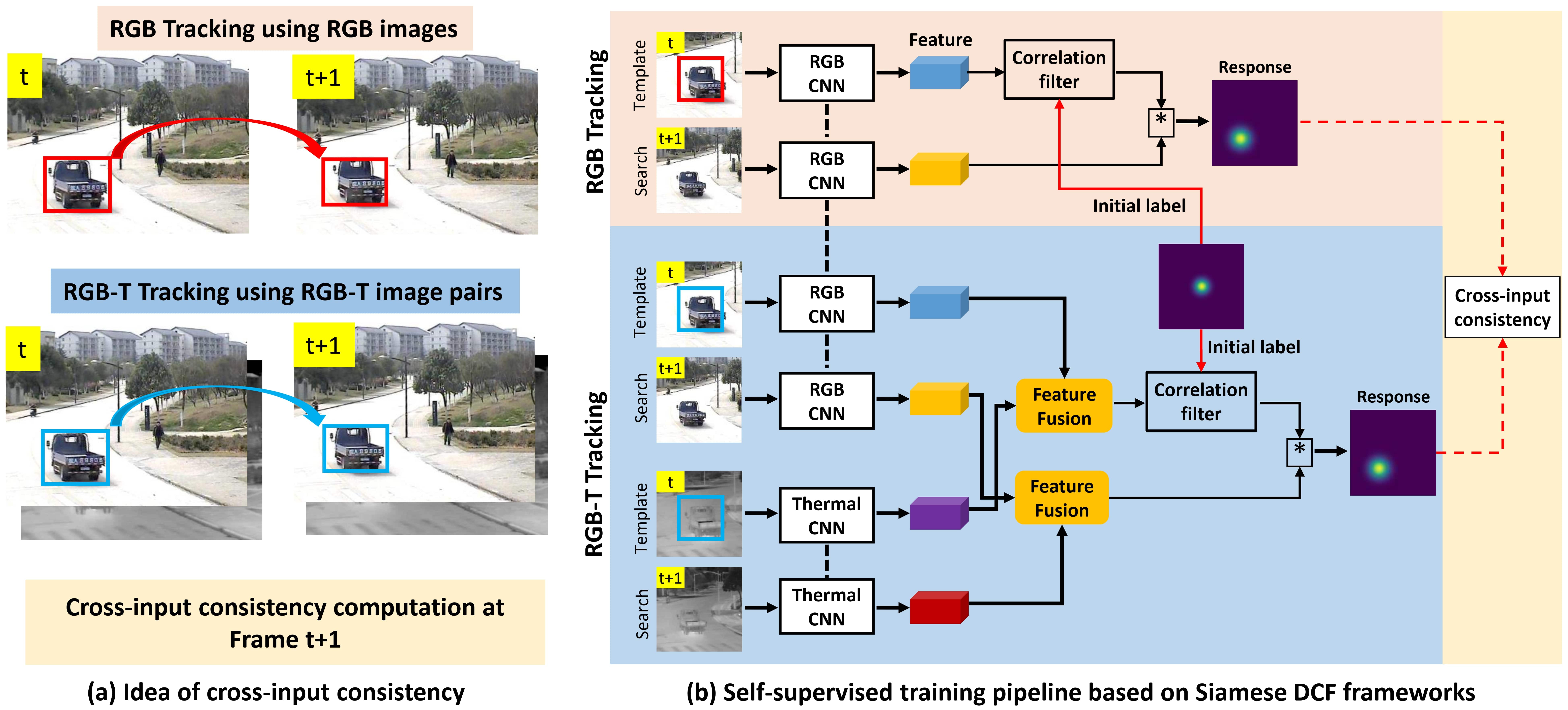}
	\caption{An overview of our method.~(a) The basic idea of cross-input consistency using RGB and RGB-T images as inputs.~(b) The training pipeline based on Siamese DCF tracking frameworks.~Loss function is computed based on response maps.~Feature-level fusion is used to obtain fused template and search features.~\textbf{The blue part in (b) is used for object tracking after training}.}
	\label{fig:idea-training}
\end{figure*}

\subsection{Self-supervised training}
Some self-supervised learning methods, e.g., BYOL \cite{grill2020bootstrap} and SimCLR \cite{chen2020simple}, first use different data augmentations to generate two correlated views and then maximize similarity to learn representations for downstream tasks.~\textit{Our idea is inspired by these self-supervised learning methods}.~However, in our work, we use
images from different modalities to replace traditional data augmentation.~Moreover, we use an object tracking framework with different inputs to exploit temporal information in RGB-T video pairs and construct cross-input consistency to guide training.~Furthermore, we do not use a pretext task and downstream tasks like many studies.~Instead, 
we only have one task (RGB-T tracking).~We directly use the proposed training method to obtain an RGB-T tracker.

\subsection{Cross-input consistency}
Cross-input consistency has been rarely utilized in tracking.~Bastani et al.~\cite{bastani2021self} applied cross-input consistency to develop a self-supervised multi-object tracker.~Our work is inspired by \cite{bastani2021self} and aims to train an RGB-T single object tracker.~We utilize RGB-T video pairs as different inputs to build cross-input consistency.

\section{Proposed Method}
\label{sec:method}
The basic idea of this work (see Fig.~\ref{fig:idea}) is that \textit{object tracking can be performed with different inputs} to generate consistent results.~Specifically, in this paper, we construct two distinct inputs, i.e., RGB images and RGB-T image pairs, to build cross-input consistency, as shown in Fig.~\ref{fig:idea-training}\xz{(a)}.~The RGB input is handled by an RGB tracker, and the RGB-T input is handled by our RGB-T tracker.~We implement our cross-input consistency self-supervised training strategy in a Siamese-based DCF tracking framework.

\subsection{Background: Siamese-based DCF tracker}
Given two consecutive frames from an unlabeled video, we first crop the template patch $\textbf{T}$ and the search patch $\textbf{S}$.~In Siamese-based DCF trackers \cite{valmadre2017end,wang2019unsupervised}, CNNs are first used to extract features from  $\textbf{T}$ and  $\textbf{S}$.~Then, a filter $\textbf{W}$ is learned, which can be used to generate a response map by convolving $\textbf{W}$ with the feature of a search patch $\textbf{S}$.~The response map is used for target localization.~Specifically, the filter $\textbf{W}$ for RGB images can be obtained as
\small
\begin{equation}
    \textbf{W}_{\rm \textbf{T}_{\rm RGB}} = \mathcal{F}^{-1}\left(\frac{\mathcal{F}(\phi_{\rm RGB}(\textbf{T}_{\rm RGB}))\odot\mathcal{F}^{\star}(\textbf{Y}_{{\rm \textbf{T}}_{\rm RGB}})}{\mathcal{F}^{\star}(\phi_{\rm RGB}(\textbf{T}_{\rm RGB})\odot\mathcal{F}(\phi_{\rm RGB}(\textbf{T}_{\rm RGB})) + \lambda}\right),
\end{equation}
\normalsize where $\odot$ is element-wise produce, $\mathcal{F}$ is the Discrete Fourier Transform (DFT), $\mathcal{F}^{-1}$ is inverse DFT, $\star$ means the complex-conjugate operation, $\phi_{\rm RGB}()$ is the CNN used to extract RGB features, $\textbf{Y}_{{\rm T}_{\rm RGB}}$ is the label of the RGB template patch, which is a Gaussian response map centered at the bounding box region.~Once the filter $ \textbf{W}_{\rm \textbf{T}_{\rm RGB}}$ is obtained, the response map of an RGB search patch $\textbf{S}_{\rm RGB}$ is
\begin{equation}
\textbf{R}_{\rm \textbf{S}_{RGB}} = \mathcal{F}^{-1}(\mathcal{F}^{\star}(\textbf{W}_{\rm \textbf{T}_{\rm RGB}})\odot \mathcal{F}(\phi_{\rm RGB}(\textbf{S}_{\rm RGB}))).
\end{equation}

The main advantage of using CNNs in DCF-based trackers is that CNNs and the CF layer are integrated into an end-to-end framework.~Therefore, the CNNs can learn to extract more suitable features for tracking.~Both our RGB tracker and RGB-T tracker use Siamese-based DCF framework, but RGB CNN and thermal CNN have different weights.

\subsection{Cross-input consistency}
As can be seen from Fig.~\ref{fig:idea-training}, our framework uses two distinct inputs to construct cross-input consistency.~The first input is RGB images, and the second input is RGB-T image pairs.~The key idea of our self-supervised training strategy is that we can arrive the location of our target in frame $(t+1)$ from frame $t$ by tracking with either input if both the RGB tracker and the RGB-T tracker work well.

In the training process, a Gaussian response map centered at the bounding box region is used as the initial label for both the RGB tracker and the RGB-T tracker.~We use a cross-input consistency loss to guide the training of the RGB tracker and the RGB-T tracker together.~The main objective is to learn the CNN models in the trackers to learn features that are suitable for tracking.~\textbf{In the inference stage, we only use the RGB-T tracker to perform tracking by using RGB and thermal images as input, as shown in the blue part in Fig.~\ref{fig:idea-training}\xz{(b)}}.

Our cross-input consistency is generic.~In this study, we use videos of different modalities to construct cross-input consistency.~There may be other schemes that can construct cross-input consistency and give comparable or better performance.~Also, as we will show in the experiments, we can also construct cross-input consistency between thermal images and RGB-T image pairs, or between RGB images, thermal images, and RGB-T image pairs.

\subsection{Our RGB-T tracker}
The architecture of our RGB-T tracker is shown in the blue part of Fig.~\ref{fig:idea-training}\xz{(b)}.~As can be seen, our RGB-T tracker consists of two RGB CNNs and two thermal CNNs.~The two RGB CNNs are used to extract RGB template and search features, and two thermal CNNs are used to extract thermal template and search features.~The RGB template feature and thermal template feature are fused to give fused template feature, while the RGB search feature and the thermal search feature are fused to give fused search feature.~Then, following \cite{wang2017dcfnet}, the fused template feature and fused search feature are used to generate response map through correlation filter and circular convolution operations, i.e.,
\begin{equation}
\small
\textbf{R}_{\rm \textbf{S}_{RGBT}} = \mathcal{F}^{-1}(\mathcal{F}^{\star}(\textbf{W}_{\rm \textbf{T}_{\rm RGBT}})\odot \mathcal{F}(\phi_{\rm RGB}(\textbf{S}_{\rm RGB})\oplus\phi_{\rm T}(\textbf{S}_{\rm T}))),
\end{equation}
where $\oplus$ means feature fusion.~Tracking result can then be obtained based on the response map.

\subsubsection{Feature fusion}
Feature fusion can be performed in various ways.~In this study, to make our RGB-T tracker lightweight so that it can run fast, we do not employ complicated feature fusion modules.~Instead, we 
concatenate the RGB feature and thermal feature to generate the fused feature.~This is simple but effective as we will show in Section \ref{subsec:ablation}.

\subsubsection{Online object tracking}
We first run offline training to train our CNNs.~Then, we perform online tracking using the RGB-T tracker.~During tracking, all CNNs are fixed.~Following previous studies \cite{wang2017dcfnet,wang2019unsupervised,wang2021unsupervised}, we update the DCF parameters in the RGB-T tracker to make the tracker more robust, i.e., 
\begin{equation}
\textbf{W}_{\rm RGBT}^t = (1-\alpha_t)\textbf{W}_{\rm RGBT}^{t-1}+\alpha_t\textbf{W}_{\rm RGBT},
\end{equation}
where $\alpha_t$ is the parameter controlling the update speed.

\subsection{Cross-input consistency loss function}
Ideally, the tracking results from different inputs should be the same if all trackers work well.~We formulate the loss function to minimize the difference between the response maps obtained using different inputs.~Specifically, our cross-input consistency loss is 
\begin{equation}
    \mathcal{L} = || \textbf{R}_{\textbf{S}_{\rm RGB}} - \textbf{R}_{\textbf{S}_{\rm RGBT}}||,
    \label{eq:loss}
\end{equation}
where $\textbf{R}_{\textbf{S}_{\rm RGB}}$ is the response map generated by the RGB tracker and $\textbf{R}_{\textbf{S}_{\rm RGBT}}$ is the response map generated by the RGB-T tracker.

\section{Training data processing and loss function re-weighting}
\label{sec:training}
\subsection{Training data processing}
\label{subsec:training-date-generation}
\begin{figure}
	\centering
	\includegraphics[width=0.45\textwidth]{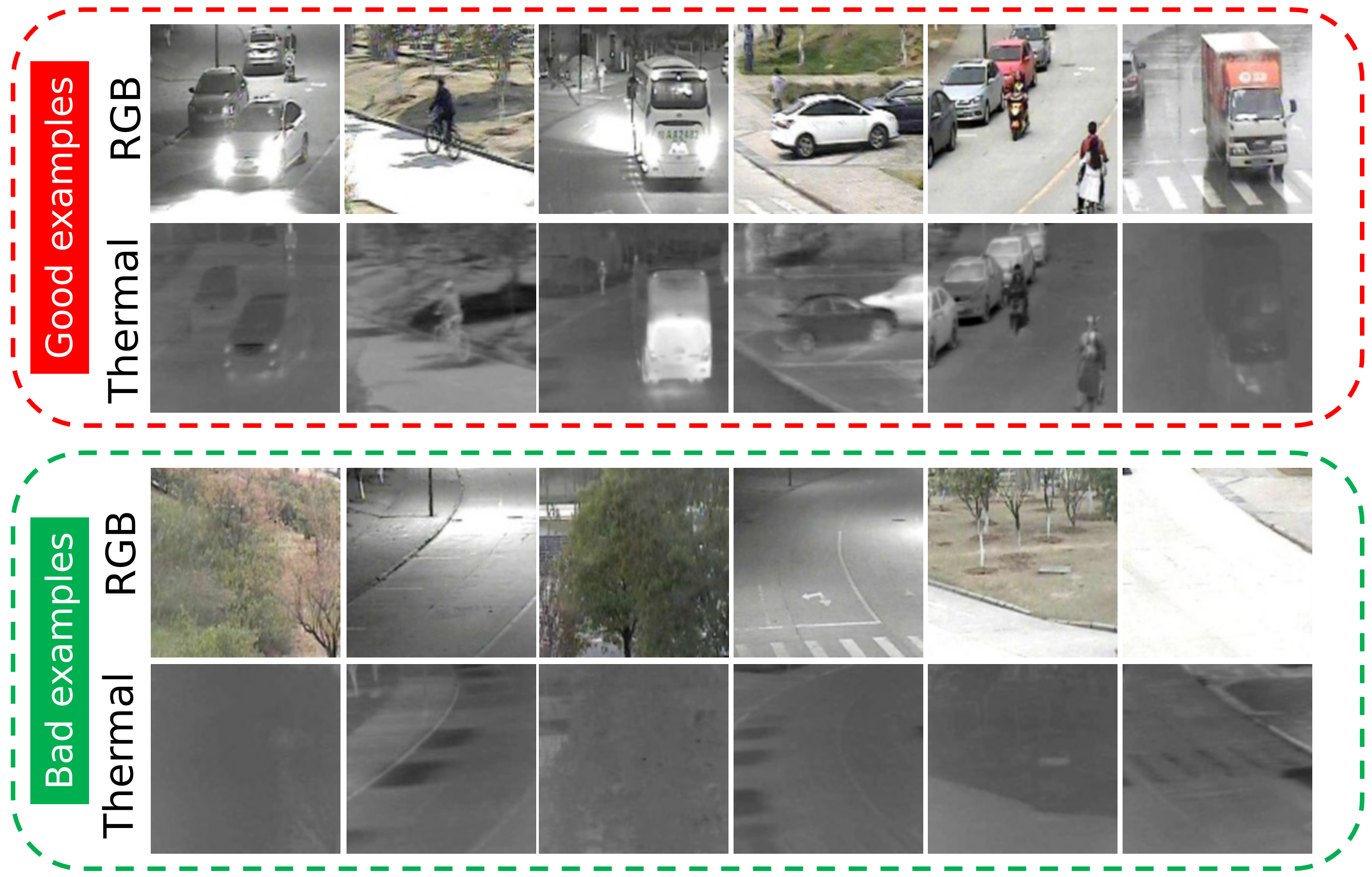}
	\caption{Examples of cropped patches from RGBT234 \cite{li2019rgb}.~Top: good examples.~Bottom: bad examples.}
	\label{fig:cropped-examples}
\end{figure}

We do not want to use any human labels in training.~It is thus essential to obtain good initial bounding boxes (pseudo labels) in self-supervised training.~In this work, we cropped the center patch from RGB-T video pairs to generate our training data, as done by Wang et al.~\cite{wang2019unsupervised}.~In this way, we track the objects appear in the center of the cropped region.~Note that the object in the center may be just a part of the object.~Some examples of the cropped images are shown in Fig.~\ref{fig:cropped-examples}.~As can be seen, some cropped images contain useful moving objects, while some images only contain background information.~In this study, we propose several ways to improve the usage of these training data, inspired by Wang et al.~\cite{wang2019unsupervised}.

\textbf{Noisy sample dropping}.~The cropped center patches contain noisy samples that provide very large loss values.~These noisy samples make the training unstable and less effective.~We assign a weight value $\textbf{D}^i_{\rm noisy}$ to each training pair to exclude 10$\%$ of train pairs that provide very high loss values.~Based on our observation, these samples usually contain sudden camera movement or sharp appearance change.~Unlike the method of Wang et al.~\cite{wang2019unsupervised} which plays with response maps, we use the difference between the RGB template patch and RGB search patch, i.e.,
\begin{equation}
    \textbf{D}^i =  \frac{|| \textbf{T}_{\rm RGB}^i - \textbf{S}_{\rm RGB}^i||^2_2}{{\rm H}\times {\rm W}},
    \label{eq:difference}
\end{equation}
where H and W are the height and width of training samples, respectively.~We 
sort the elements in $\textbf{D}$.~Then, we assign a weight value $\textbf{D}^i_{\rm noisy}$ to each training pair.~$10\%$ of elements in the weight vector $\textbf{D}_{\rm noisy}$ corresponding to noisy samples are 0.~In this way, we exclude $10\%$ of training pairs that produce large difference values.

\textbf{Background sample dropping}.~As shown in 
Fig.~\ref{fig:cropped-examples}, some cropped center patches contain only background or still objects.~These background samples make little contribution to model training.~To exclude these background 
training samples, we set a value $\textbf{D}^i_{\rm background}$ to each training pair.~$25\%$ of the elements in the weight vector $\textbf{D}_{\rm background}$ corresponding to the $25\%$ lowest values in $\textbf{D}$ are zero.~Combining $\textbf{D}_{\rm noisy}$ and $\textbf{D}_{\rm background}$, we can normalize the weight of each training pair to ensure the sum of useful weights in one mini-batch is 1, i.e.,
\begin{equation}
    \textbf{D}_{\rm norm}^i = \frac{\textbf{D}^i_{\rm noisy}\cdot \textbf{D}^i_{\rm background}}{\sum_{i=1}^n\textbf{D}^i_{\rm noisy}\cdot \textbf{D}^i_{\rm background}},
    \label{eq:weight-vector}
\end{equation}
where $n$ is the number of training pairs in a mini-batch.

\subsection{Loss function re-weighting}
After computing the re-weighting weight vector using Eq.~(\ref{eq:weight-vector}), we use it to re-weight the loss obtained from training samples of various quality, i.e.,
\begin{equation}
    \mathcal{L}_{\rm final} = \frac{1}{n}\sum_{i=1}^n    \textbf{D}_{\rm norm}^i \cdot \mathcal{L}^i,
\end{equation}
where $\mathcal{L}^i$ is computed using Eq.~(\ref{eq:loss}).~Using this loss can make the training more effective and avoid overfitting.

\section{Experiments}
\label{sec:experiments}
\textbf{Implementation details}.~Following \cite{wang2019unsupervised, wang2021unsupervised}, we use  lightweight CNNs in our trackers.~Specifically, the filter sizes of the two convolutional layers in our 
CNN are $3\times3\times3\times32$ and $3\times3\times32\times32$.~All experiments were performed using a desktop equipped with two NVIDIA RTX3090 GPUs and an i9-10900X CPU.~The batch size is 32.~We change the learning rate from $10^{-4}$ to $10^{-6}$ from epoch 0 to epoch 30.~The weight decay is $5\times10^{-5}$.

\textbf{Test set}.~We report results on the GTOT dataset \cite{li2016learning} and the RGBT234 dataset, which have been widely used in RGB-T tracking studies \cite{li2019learning,wang2020cross,zhang2022visible}.~GTOT consists of 50 RGB-T videos (15.8K frames).~Moreover, seven attributes are annotated for each sequence, including occlusion (OCC), large scale variation (LSV), fast motion (FM), low illumination (LI), thermal crossover (TC), small object (SO),  and deformation (DEF).~RGBT234 contains 234 RGB-T video pairs (around 233.8K frames) and 12 attributes are annotated.~Compared to GTOT, RGBT234 is more challenging by having longer frames in videos and more  challenging attributes.

\textbf{Training data}.~When testing on the GTOT dataset, we use the RGBT234 dataset \cite{li2019rgb} as training data.~10,000 RGB-T pairs are randomly chosen as the validation set in training.~When testing on the RGBT234 dataset, we use the GTOT dataset as training data, and 1000 RGB-T pairs are randomly chosen as the validation set in training.

\textbf{Evaluation metrics}.~In this work, we utilize two commonly-used evaluation metrics in RGB-T tracking, maximum precision rate (MPR) and maximum success rate (MSR) \cite{li2019rgb,zhang2021learning}, to evaluate the performance of our tracker.~Following previous studies \cite{li2016learning,li2019rgb,zhang2021jointly}, the threshold of MPR is set to 5 pixels for GTOT (because the targets in GTOT are relatively small) and 20 pixels for RGBT234.

\subsection{Self-supervised v.s.~supervised training}
To show the effectiveness of our self-supervised training strategy, we use the ground truth of the RGBT234 dataset to train a supervised RGB-T tracker.~Specifically, we only train the RGB-T tracker shown in the blue part of Fig.~\ref{fig:idea-training}\xz{(b)}.~The comparison between the supervised RGB-T tracker and our self-supervised RGB-T tracker is shown in Table \ref{table:results-supervised}.~As can be seen, our self-supervised RGB-T tracker achieves better performance than the supervised one on GTOT.~This is interesting and supervising, as training using ground truth labels is usually more effective.~A possible reason is that by using center-cropped regions from RGBT234 (contains 110K RGB-T image pairs) as training data, the training set has more categories of targets than the ground truth labels.~Similar pattern has been observed in some unsupervised RGB tracking studies \cite{yang2019learning}, where the unsupervised DCFNet performs slightly better than supervised DCFNet.~We also use the ground truth of the GTOT dataset to train a supervised RGB-T tracker an test it on RGBT234.~As can be seen from Table \ref{table:results-supervised}, our self-supervised RGB-T tracker is slightly worse than its supervised counterpart.~This may because the GTOT dataset is smaller (7.9K RGB-T image pairs) 
and does not provide enough high-quality training data for our self-supervised training.

\begin{table}[H]
	\caption{Comparison of the proposed self-supervised training and supervised training.~Better results are maked in bold.}
	\label{table:results-supervised}
	\centering
	\resizebox{0.4\textwidth}{!}{
	\begin{tabular}{|c|cc|cc|}
		\hline 
    \multirow{2}{*}{Variant} & 	\multicolumn{2}{c|}{GTOT} & 	\multicolumn{2}{c|}{RGBT234} \\ \cline{2-5}
    & MPR($\uparrow$) & MSR($\uparrow$)  & MPR($\uparrow$) & MSR($\uparrow$)   \\   \cline{1-5} 
    Supervised & 76.7 & 66.1   &\textbf{59.5}& \textbf{43.7} \\
    Ours &\textbf{85.6} &\textbf{70.5} & 56.2 & 41.6\\ \hline 
	\end{tabular}}
\end{table}

\subsection{Ablation studies and analysis}
\label{subsec:ablation}
The GTOT dataset is used in ablation studies 
unless otherwise specified.

\vspace{0.05cm}
\noindent\textbf{Different cross input combinations}.~We can use different combinations of inputs to construct cross-input consistency.~In this study, we keep the second input (RGB-T image pairs) and change the first input to different variants.~Specifically, we trained two variants.~In the first variant, we use thermal images as the first input.~In the second variant, we use 4-channel RGB-T images as the first input.~When using 4-channel RGB-T images, we change the dimension of the first convolution layer to adapt to these 4-channel images.~As shown in Table \ref{table:results-first-input}, all these combinations can be used to guide our cross-input consistency-based training and gives useful RGB-T trackers.~Among these combinations, when the first input is RGB image and the second input is RGB-T image pairs, our RGB-T tracker shows the best tracking performance.
\begin{table}
	\caption{Effect of input in the first branch.}
	\label{table:results-first-input}
	\centering
	\resizebox{0.45\textwidth}{!}{
		\begin{tabular}{|c|c|cc|}
			\hline 
			First input & Second input 	& MPR($\uparrow$) & MSR($\uparrow$)  \\\cline{1-4}   
		 	RGB  & RGB-T &\textbf{85.6} &\textbf{70.5 } \\ 
			Thermal & RGB-T   &75.4 &64.9     \\ 
			4-channel RGBT  &RGB-T & 81.6  &67.8   \\ 
			\hline 
		\end{tabular}
	}
\end{table}

\vspace{0.05cm}
\noindent\textbf{Using more branches in cross-input consistency}.~Previously, we used two branches (different inputs) to build cross-input consistency, as shown in Fig.~\ref{fig:idea-training}.~We can extend our idea to use more branches.~For example, we can use three inputs, i.e., RGB images, thermal images, and RGB-T image pairs.~In this case, we can add another two loss terms, i.e., $\mathcal{L}_{\rm RGB-T}$ and $\mathcal{L}_{\rm T-RGBT}$, to compute the difference of every two outputs.~Using this idea, we obtain an RGB-T tracker with MPR of 83.7 and MSR of 70.3, which is better than using 4-channel RGBT images and RGB-T image pairs.~However, the performance is slightly worse than using RGB images and RGB-T image pairs as distinct inputs.

\noindent\textbf{Impact of loss function}.~Mean square error (MSE) loss is usually used to train unsupervised trackers \cite{wang2019unsupervised, wang2021unsupervised,huang2022thermal}.~In this study, we use L1 loss instead.~Table \ref{table:impact-loss} shows the performance comparison of using MSE loss and L1 loss in our cross-input consistency-based self-supervised training.~As can be seen, L1 loss gives better performance in all cases.~This may because our training samples are noisy (although we use re-weighting strategy), MSE loss will amplify the errors due to noisy training samples, making the training less effective.
\begin{table}
	\caption{Impact of loss function.~In these variants, L1 loss gives better performance than MSE loss.}
	\label{table:impact-loss}
	\centering
	\resizebox{0.4\textwidth}{!}{
		\begin{tabular}{|c|c|c|c|}
			\hline 
\multicolumn{2}{|c|}{Variant} & MPR($\uparrow$) & MSR($\uparrow$)  \\\cline{1-4}     
\multirow{2}{*}{Three branches} &	 MSE loss  &72.5  &63.1  \\ 
&	L1 loss   &75.3  &65.6    \\ \cline{1-4} 
\multirow{2}{*}{RGB, RGB-T}  &MSE loss &78.8 &64.9 \\
	&L1 loss &\textbf{85.6} &\textbf{70.5} \\ \cline{1-4} 
	\multirow{2}{*}{Thermal, RGB-T} & 	MSE loss & 65.7&58.2 \\
& L1 loss & 75.4&64.9 \\ 
			\hline 
		\end{tabular}
	}
\end{table}

\noindent\textbf{Impact of loss re-weighting}.~We proposed two components to generate weight vectors based on cropped patches, i.e., noisy sample dropping and background sample dropping.~In this section, we report the results of removing one of the components in Table \ref{table:results-data}.~As can be seen, after we remove any comment, the tracking performance will drop slightly, showing that our loss re-weighting strategy is helpful.

\begin{table}
	\caption{Impact of loss re-weighting scheme.~Better results are obtained after re-weighting the loss function.}
	\label{table:results-data}
	\centering
	\resizebox{0.42\textwidth}{!}{
		\begin{tabular}{|cc|cc|}
			\hline 
			\multicolumn{2}{|c|}{Loss re-weighting} 	& \multirow{2}{*}{MPR($\uparrow$)} & \multirow{2}{*}{MSR($\uparrow$)}  \\\cline{1-2} 
			\tabincell{c}{\tabincell{c}{Noisy sample\\ dropping}} & \tabincell{c}{Background sample\\ dropping}  &  &  \\ 		\hline 
			  &  & 82.6& 69.1  \\ 
			 & \Checkmark  &83.7 & 69.5    \\ 
			\Checkmark  & \Checkmark   & \textbf{85.6}  & \textbf{70.5}  \\ 
			\hline 
		\end{tabular}
	}
\end{table}

\noindent\textbf{Impact of unlabeled training data size}.~We use different portions of RGBT234 as the training set.
~As can be seen, in general, the proposed self-supervised training strategy benefits from training using more unlabeled RGB-T video pairs.~Because unlabeled RGB-T pairs are much easier to obtain than annotated ones, our method infers the great potential of unsupervised RGB-T tracking.

\begin{table}[h]
	\caption{Ablation studies on training data size.~With more unlabeled RGB-T videos for training, the proposed RGB-T tracker achieves better results on the GTOT dataset.}
	\label{table:results-training-size}
	\centering
	\resizebox{0.48\textwidth}{!}{
	\begin{tabular}{|c|cc|c|cc|}
		\hline  
		Size	&  MPR($\uparrow$) & MSR($\uparrow$) & Size	&  MPR($\uparrow$) & MSR($\uparrow$)  \\   
		\hline 
		RGBT234 (90\%)    & \textbf{85.6} & \textbf{70.5} & RGBT234 (50\%)    &81.5 &67.5  \\ 
		RGBT234 (70\%)    & 81.2 & 67.3 & RGBT234 (20$\%$)   &73.6  &61.3  \\
		\hline 
	\end{tabular}
	}
\end{table}

\vspace{0.08cm}
\noindent\textbf{Impact of feature fusion methods}.~We trained three variants of RGB-T trackers using three feature-level fusion methods, i.e., element-wise average, concatenation, and the DFF fusion module proposed by Zhang et al.~\cite{zhang2022visible}.~Specifically, we keep the RGB tracker in Fig.~\ref{fig:idea-training}\xz{(b)} 
and change the feature fusion method in the RGB-T tracker.~We compare different fusion methods in Table \ref{table:results-different-fusion}, which shows the fusion level has a significant impact on the performance of RGB-T trackers.~Specifically, by concatenating RGB features and thermal features, we obtain the best tracking performance.

\begin{table}[h]
	\caption{Impact of feature fusion methods.~Concatenation is simple yet gives the best performance.}
	\label{table:results-different-fusion}
	\centering
	\resizebox{0.35\textwidth}{!}{
	\begin{tabular}{|c|cc|}
		\hline  
		Variant	&  MPR($\uparrow$) & MSR($\uparrow$)  \\   
		\hline 
		Feature-level (Average) & 83.7 & 69.1  \\
		Feature-level (Concat.)  & \textbf{85.6} & \textbf{70.5 }   \\
		Feature-level (DFF \cite{zhang2022visible}) &79.1 &66.0 \\
		\hline 
	\end{tabular}
	}
\end{table}

\vspace{0.08cm}
\noindent\textbf{Impact of sharing weights between RGB CNN and thermal CNN}.~In our experiments, we find that whether the weights are shared between RGB CNN and thermal CNN (please see Fig.~\xz{3(b)} of the paper) or not affects the performance of our tracker.~We designed a variant of our method, i.e., the weights of the RGB CNN are shared with the thermal CNN.~The results are shown in Table \ref{table:results-shared-weights}.~As can be seen, our RGB-T tracker gives better performance than the variant where the weights are shared between the RGB CNN and the thermal CNN.

\begin{table}[H]
	\caption{Impact of sharing weights between RGB CNN and thermal CNN on tracking performance.~The GTOT dataset is used.}
	\label{table:results-shared-weights}
	\centering
	\resizebox{0.3\textwidth}{!}{
	\begin{tabular}{|c|cc|}
		\hline  
Variant	&	 MPR($\uparrow$) & MSR($\uparrow$)  \\   
		\hline 
Sharing & 82.2 & 68.2 \\
Not Sharing&\textbf{85.6} & \textbf{70.5} \\  
		\hline 
	\end{tabular}
	}
\end{table}

\begin{figure}
	\centering
	\includegraphics[width=0.48\textwidth]{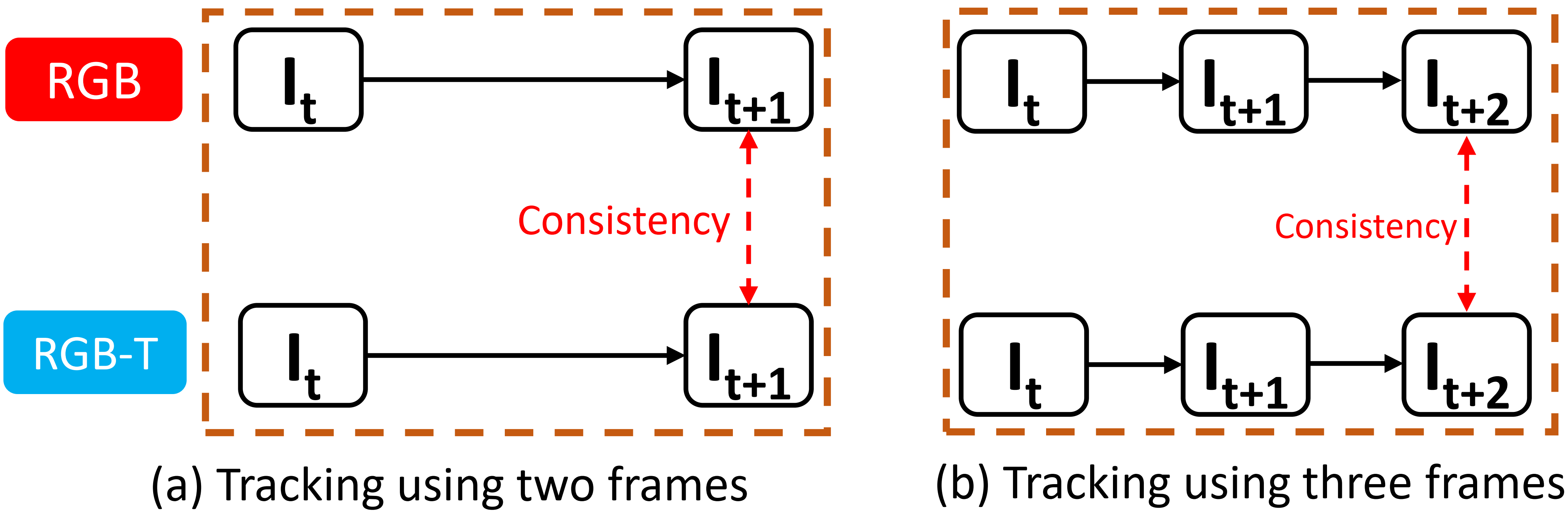}
	\caption{Training using two frames (a) and three frames (b).~\xz{Red arrow}: the cross-input consistency.}
	\label{fig:tracking-length}
\end{figure}

\vspace{0.08cm}
\noindent\textbf{Impact of tracking sequence length}.~In this paper, we build our cross-input consistency-based training strategy using two frames, i.e., frame $t$ and frame $(t+1)$, as shown in Fig.~\ref{fig:tracking-length}\xz{(a)}.~Indeed, our cross-input consistency-based training strategy can be extended to more frames, e.g., frames $t$, $(t+1)$ and $(t+2)$, as shown in Fig.~\ref{fig:tracking-length}.~In this case, the response map generated at frame $(t+1)$ will be used as the pseudo label of frame $(t+1)$.~Based on the pseudo label, our tracker tracks the target from $(t+1)$ to frame $(t+2)$.

When more frames are used, the cross-input consistency loss is computed in the final frame, as shown by the red dash arrow in Fig.~\ref{fig:tracking-length}\xz{(b)}.~We show the impact of tracking sequence length on the tracking performance in Table~\ref{table:length}.~From the results, we can see that using three frames can also train a good RGB-T tracker.~However, the performance is slightly worse than using two frames in training.~These results indicate that our self-supervised training strategy can train our RGB-T tracker effectively using only two frames in each training pair.

\begin{table}[H]
	\caption{Impact of tracking sequence length on tracking performance.~The GTOT dataset is used.}
	\label{table:length}
	\centering
	\resizebox{0.35\textwidth}{!}{
	\begin{tabular}{|c|cc|}
		\hline  
Variant	&	 MPR($\uparrow$) & MSR($\uparrow$)  \\   
		\hline 
		Using three frames & 84.3 & 69.5 \\
Using two frames & \textbf{85.6} & \textbf{70.5} \\  
		\hline 
	\end{tabular}
	}
\end{table}

\subsection{Compared with cycle consistency}
\begin{figure}
	\centering
	\includegraphics[width=0.45\textwidth]{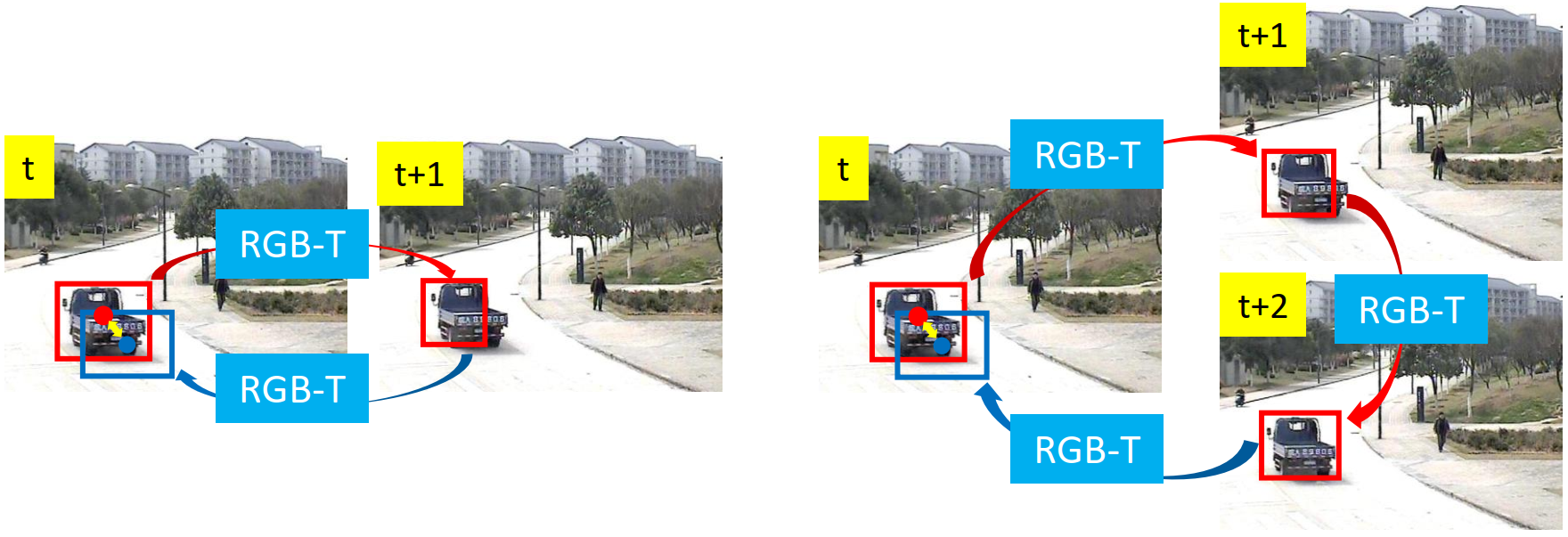}
	\caption{Cycle consistency using two frames (left) and three frames (right).}
	\label{fig:cycle}
\end{figure}

Cycle consistency is commonly used in training unsupervised RGB-based trackers \cite{wang2019unsupervised, yuan2020self,yuan2021self,wang2021unsupervised,zhu2021contrastive,shen2022unsupervised} or thermal-based trackers \cite{huang2022thermal}.~In this section, we compare our cross-input consistency with 
cycle consistency.~Specifically, we train our RGB-T tracker (the blue part of Fig.~\ref{fig:idea-training}\xz{(b)}) using cycle consistency.~We trained two variants, one with two frames and one with three frames, as shown in Fig.~\ref{fig:cycle}.~All other settings are kept the same as our cross-input consistency-based training.~As can be seen from Table \ref{table:comparison-others}, our cross-input consistency-based self-supervised training strategy is more effective than cycle consistency.~The attributed-based performance given in Table \ref{table:attribute-based} also indicates that our cross-input consistency-based training is more effective than cycle consistency-based training.~Moreover, using three frames in cycle consistency-based training is more effective than using two frames.~In addition to the RGB-T tracker, we also use our trained CNNs to run an RGB tracker and a thermal tracker based on the red part of Fig.~\ref{fig:idea-training}\xz{(b)}.~As can be seen, our training strategy provides consistent better performance in RGB tracker, thermal tracker and RGB-T tracker.~The reason is that the forward-backward tracking-based cycle consistency may not hold in some challenging real world tracking scenarios \cite{li2021self}.
~In addition, it should be mentioned that the cycle consistency-based strategy, e.g., UDT \cite{wang2019unsupervised}, has only been applied to uni-modal object tracking, while the proposed cross-input consistency-based self-supervised training strategy is applied to multi-modal tracking. 

\begin{table}[H]
	\caption{Performance comparison with cycle consistency.}
	\label{table:comparison-others}
	\centering
	\resizebox{0.48\textwidth}{!}{
	\begin{tabular}{|c|cc|cc|cc|}
		\hline  
	\multirow{2}{*}{Variant} & \multicolumn{2}{c|}{RGB tracker}  &  \multicolumn{2}{c|}{T tracker} &  \multicolumn{2}{c|}{RGB-T tracker} \\  \cline{2-7}
	&	PR & SR &  PR & SR & MPR & MSR  \\   
		\hline
     Cycle consistency (2 frames) & 67.6 &56.7  &62.4 &56.0 &  74.1& 64.5 \\ 
	 Cycle consistency (3 frames)  &64.5 &55.6  &62.8 &56.2& 75.4 &65.6  \\ 
     Ours &\textbf{68.1}&\textbf{57.8}  &\textbf{64.8} &\textbf{57.3}  &\textbf{85.6} &\textbf{70.5} \\ \hline 
	\end{tabular}
	}
\end{table}

\begin{table*}[t]
	\caption{Comparison with existing RGB-T trackers on GTOT dataset and RGBT234 dataset.~The results marked with $\dag$ are computed by us using raw tracking results.~The results marked with $\S$ are copied from \cite{li2019rgb,li2016learning,li2021rgbt}.~Other results are extracted from corresponding papers.~`-' means not mentioned in the corresponding paper.~Values worse than our method are marked in \colorbox{green!30}{\color{black}green}.}
	\label{table:comparison-with-RGBT}
	\centering
	\resizebox{0.98\textwidth}{!}{
	\begin{tabular}{|c|cc|cc|c|c|c|c|}
		\hline 
		\multirow{2}{*}{Tracker} &  \multicolumn{2}{c|}{GTOT}  & \multicolumn{2}{c|}{RGBT234} & \multirow{2}{*}{FPS ($\uparrow$)} & \multirow{2}{*}{Category}  & \multirow{2}{*}{\tabincell{c}{Supervised}} & \multirow{2}{*}{Venue} \\ \cline{2-5} 
			& MPR ($\uparrow$)& MSR ($\uparrow$) & MPR ($\uparrow$) & MSR ($\uparrow$) &  & & & \\   
		\hline 
	    HMFT$\dag$ \cite{zhang2022visible} & 90.6  & 74.2 & 78.8 & 56.8  & \cellcolor{green!30}30.2 & DL-based &\cellcolor{green!30}Yes  & CVPR 2022   \\
		CMPP \cite{wang2020cross} & 92.6& 73.8 & 82.3 & 57.5 & \cellcolor{green!30}1.3 & DL-based & \cellcolor{green!30}Yes & CVPR 2020 \\
		APFNet \cite{xiao2022attribute} & 90.5 & 73.7 & 82.7& 57.9& - & DL-based & \cellcolor{green!30}Yes & AAAI 2022  \\ 
		DMCNet \cite{lu2022duality} &90.9 &73.3 & 83.9 &59.3 & \cellcolor{green!30}2.4 & DL-based & \cellcolor{green!30}Yes  & IEEE TNNLS 2022 \\
		JMMAC$\dag$ \cite{zhang2021jointly}& 89.3& 73.1 & 79.0& 57.3 & \cellcolor{green!30}4 & DL-based & \cellcolor{green!30}Yes  & IEEE TIP 2021\\
		CBPNet \cite{xu2021multimodal} & 88.5 & 71.6 &79.4 & 54.1 & \cellcolor{green!30}3.7 & DL-based & \cellcolor{green!30}Yes  & IEEE TMM 2021 \\ 
		MaCNet \cite{zhang2020object} &88.0 &71.4  & 79.0& 55.4 & \cellcolor{green!30}0.8 & DL-based & \cellcolor{green!30}Yes& Sensors 2020 \\
		MANet$\dag$ \cite{li2019multi}&88.9 & 71.1 &77.7 &53.9  & \cellcolor{green!30}3.1& DL-based & \cellcolor{green!30}Yes  & ICCVW 2019 \\	
		CMP \cite{yang2021rgbt} & 86.9 & 71.1 & 75.1 & 49.1 & \cellcolor{green!30}35 & DL-based &  \cellcolor{green!30}Yes & Neurocomputing 2021  \\
		MFGNet \cite{wang2022mfgnet} & 88.9 & 70.7  & 78.3 & 53.5 & \cellcolor{green!30}3.4 & DL-based & \cellcolor{green!30}Yes & IEEE TMM 2022\\ \hline 
		DAFNet$\dag$ \cite{gao2019deep} & 88.6 &  \cellcolor{green!30}69.9 & 79.6 & 54.4   & \cellcolor{green!30}23 & DL-based & \cellcolor{green!30}Yes& ICCVW 2019 \\ 
		DAPNet$\dag$ \cite{zhu2019dense}&87.4 & \cellcolor{green!30}68.9 & 76.6& 53.7  & \cellcolor{green!30}2& DL-based & \cellcolor{green!30}Yes & ACM MM 2019\\
		mfDiMP$\S$  \cite{zhang2019multi}&\cellcolor{green!30}84.1 & \cellcolor{green!30}69.3  &78.5 &55.9  & \cellcolor{green!30}10.3& DL-based& \cellcolor{green!30}Yes & ICCVW 2019 \\ 
		LTDA \cite{yang2019learning} &\cellcolor{green!30}84.3 & \cellcolor{green!30}67.7 &78.7 &54.5  & \cellcolor{green!30}0.4& DL-based & \cellcolor{green!30}Yes & ICIP 2019 \\ 	
		DuSiamRT \cite{guo2022dual} & \cellcolor{green!30}76.6 & \cellcolor{green!30}62.8  &  56.7 &  \cellcolor{green!30}38.4 & \cellcolor{green!30}116 & DL-based & \cellcolor{green!30}Yes & The Visual Computer 2022\\
		TCNN \cite{li2018fusing} & \cellcolor{green!30}85.2 & \cellcolor{green!30}62.6 &- &- & \cellcolor{green!30}15 & DL-based& \cellcolor{green!30}Yes  & Neurocomputing 2018\\
		JCDA-InvSR \cite{kang2019grayscale} & -  &\cellcolor{green!30}60.5 & 60.6 &  \cellcolor{green!30}41.4  & \cellcolor{green!30}1.6 & ML-based & \cellcolor{green!30}Yes& IEEE TIP 2019\\ 
		
		CFNet \cite{valmadre2017end} + RGBT$\S$ & - & - & \cellcolor{green!30}55.1 & \cellcolor{green!30}39.0 & - & DL-based &  \cellcolor{green!30}Yes & CVPR 2017 \\
		SiamDW \cite{zhang2019deeper} + RGBT$\S$ & \cellcolor{green!30}68.0 & \cellcolor{green!30}56.5& 60.4 & \cellcolor{green!30}39.7 &- &DL-based & \cellcolor{green!30}Yes &CVPR 2019  \\
		
		\hline 
		CMCF \cite{zhai2019fast} & \cellcolor{green!30}77.0 & \cellcolor{green!30}63.2 & - & - & 227 & Non-DL (CF-based) &  & Neurocomputing 2019 \\
		NRCMR \cite{li2021rgbt} & \cellcolor{green!30}83.7 & \cellcolor{green!30}66.4 & 72.9 & 50.2  & \cellcolor{green!30}7&Non-DL (Graph-based) &  & IEEE TNNLS 2021\\
		LGMG \cite{li2019learning} & \cellcolor{green!30}83.7 & \cellcolor{green!30}65.8 & -&-  &\cellcolor{green!30}7 & Non-DL (Graph-based)&  & IEEE TCSVT 2019\\
		CMR \cite{li2018cross} & \cellcolor{green!30}82.7 & \cellcolor{green!30}64.3  & - & -  &\cellcolor{green!30}8 & Non-DL (Graph-based)& & ECCV 2018\\

		SGT$\S$   \cite{li2017weighted}& \cellcolor{green!30}85.1 & \cellcolor{green!30}62.8  &72.0 & 47.2 & \cellcolor{green!30}5 &Non-DL (Graph-based) &  & ACM MM 2017\\ 
		\cite{li2018two} & \cellcolor{green!30}84.2 & \cellcolor{green!30}62.2  & - &- &\cellcolor{green!30}7 &Non-DL (Graph-based) &  & SPIC 2018 \\ 
		CSR \cite{li2016learning} & \cellcolor{green!30}74.5 & \cellcolor{green!30}61.5  &  \cellcolor{green!30}46.3&  \cellcolor{green!30}32.8 & \cellcolor{green!30}1.6&Non-DL (SR-based) & &IEEE TIP 2016 \\
		\cite{shen2022rgbt} & \cellcolor{green!30}77.3 & \cellcolor{green!30}61.2 &72.9 &48.6 &\cellcolor{green!30}1.2 & Non-DL (Graph-based)&  & Neurocomputing 2022\\ 
		
		MEEF\cite{zhang2014meem} + RGBT$\S$ &- & \cellcolor{green!30}52.0& 63.6 & \cellcolor{green!30}40.5 & \cellcolor{green!30}4.9 & Non-DL &   &ECCV 2014 \\  
		
		KCF \cite{henriques2014high} + RGBT$\S$ &- & \cellcolor{green!30}42.0& \cellcolor{green!30}46.3 & \cellcolor{green!30}30.5 & \cellcolor{green!30}124.1 & Non-DL (CF-based) &  & IEEE TPAMI 2014   \\ 
		
		\hline 
		
		\cellcolor{gray!40}Ours & \cellcolor{gray!40}85.6 & \cellcolor{gray!40}70.5 &\cellcolor{gray!40}56.2 &\cellcolor{gray!40}41.6 & \cellcolor{gray!40}179.3 & \cellcolor{gray!40}DL-based & \cellcolor{gray!40}No &\cellcolor{gray!40}\\
	\hline 
	\end{tabular}
	}
\end{table*}

\subsection{Compared with SOTA RGB-T tackers}
\noindent\textbf{Compared methods}.~There are no existing unsupervised deep RGB-T trackers.~We selected the following supervised RGB-T trackers for comparison, i.e., HMFT \cite{zhang2022visible}, CMPP \cite{wang2020cross}, DMCNet \cite{lu2022duality}, JMMAC \cite{zhang2021jointly}, CBPNet \cite{xu2021multimodal}, MaCNet \cite{zhang2020object}, MANet \cite{li2019multi}, CMP \cite{yang2021rgbt}, MFGNet \cite{wang2022mfgnet}, DAFNet \cite{gao2019deep},  DAPNet \cite{zhu2019dense}, mfDiMP \cite{zhang2019multi}, LTDA \cite{yang2019learning}, DuSiamRT \cite{guo2022dual}, TCNN \cite{li2018fusing}, JCDA-InvSR \cite{kang2019grayscale}, CFNet \cite{valmadre2017end}+RGBT, SiamDW \cite{zhang2019deeper}+RGBT.~A transformer-based method, namely APFNet \cite{xiao2022attribute}, is also selected for comparison.~We also selected some non-deep RGB-T trackers, including CMCF \cite{zhai2019fast}, NRCMR \cite{li2021rgbt}, LGMG \cite{li2019learning}, CMR \cite{li2018cross},  SGT \cite{li2017weighted}, the method of Li et al.~\cite{li2018two}, CSR \cite{li2016learning}, \cite{shen2022rgbt}, MEEF\cite{zhang2014meem}+RGBT and KCF \cite{henriques2014high}+RGBT.~Almost all categories of RGB-T methods are covered.

\noindent\textbf{Results on GTOT}.~The tracking results on the GTOT dataset are shown in Table \ref{table:comparison-with-RGBT}.~As can be seen, our RGB-T tracker is better than all non-learning-based RGB-T trackers in terms of both metrics.~Furthermore, our self-supervised RGB-T tracker is better than seven supervised RGB-T trackers (DAFNet, DAPNet, mfDiMP, LTDA, DuSiamRT, TCNN, JCDA-InvSR) and 
the RGB-T version of CFNet.~Our tracker also shows comparable performance with several supervised trackers, such as MFGNet, CMP and MANet.~Although there are some gaps between our performance and the state-of-the-art supervised RGB-T trackers, it is understandable because 
those supervised RGB-T trackers use large-scale annotated RGB-T image pairs for training.~In contrast, our method does not use any annotations.~Moreover, those trackers use more complex models, as indicated by their tracking speed.~For example, the FPS of CMPP and DMCNet are 1.3 and 2.4, respectively.

\textbf{Attribute-based results}.~Attribute-based performance on the GTOT dataset is shown in Table \ref{table:attribute-based}.~As can be seen, as a very lightweight RGB-T tracker trained without any ground truth labels, our tracker achieves very competitive performance in terms of LSV, LI, TC and SO.~Especially, our RGB-T tracker achieves better performance than most supervised RGB-T trackers in terms of LSV, indicating that our tracker can well handle scale variation of targets.

\begin{table}
	\begin{center}
		\caption{Attribute-based performance (MSR) on GTOT.~Values worse than our method are marked in \colorbox{green!30}{\color{black}green}.}
		\label{table:attribute-based}
		\small
		\resizebox{0.45\textwidth}{!}{
		\begin{tabular}{|c|c|c|c|c|c|c|c|}	
			\hline
			Method
	        & OCC  & LSV  & FM & LI & TC & SO & DEF    \\  \hline
	        HMFT & 72.1 & 74.0 &75.2 &75.6 &72.3 &70.8 &72.3\\
	        MANet & 68.9& \cellcolor{green!30}69.4 &69.2 & \cellcolor{green!30}71.9 &69.1&68.6&73.4\\
	        CBPNet &68.6 &\cellcolor{green!30}68.8 & 67.5 & 73.1 & 69.3 & 69.0 &75.4 \\
	        MaCNet & 68.7 & \cellcolor{green!30}67.3 & 65.9 &73.1 &69.7 &69.5 &76.5  \\
	        DAPNet & 67.4 &\cellcolor{green!30}64.8 & \cellcolor{green!30}61.9 & \cellcolor{green!30}72.2 &69.0 & 69.2 & 77.1 \\ 
	        DuSiamRT & \cellcolor{green!30}57.7 &\cellcolor{green!30}64.5 &\cellcolor{green!30}58.0 & \cellcolor{green!30}62.3 & \cellcolor{green!30}61.4 & \cellcolor{green!30}64.2 & \cellcolor{green!30}62.9 \\
	        TCNN & \cellcolor{green!30}55.6 & \cellcolor{green!30}64.6 & \cellcolor{green!30}51.7 & \cellcolor{green!30}64.2  &\cellcolor{green!30}59.5 & \cellcolor{green!30}59.1 & 73.4 \\
	        SGT& \cellcolor{green!30}56.7 & \cellcolor{green!30}54.7 & \cellcolor{green!30}55.9 &\cellcolor{green!30}65.1 &\cellcolor{green!30}61.5&\cellcolor{green!30}61.8 &73.3 \\\hline 
	        Cycle-2 &\cellcolor{green!30}59.1 &\cellcolor{green!30}67.4  & \cellcolor{green!30}60.3 & \cellcolor{green!30}63.9 & \cellcolor{green!30}65.7 &\cellcolor{green!30}61.9  & \cellcolor{green!30}61.3  \\	
	        Cycle-3 &\cellcolor{green!30}62.0 & \cellcolor{green!30}67.1 & \cellcolor{green!30}62.9&\cellcolor{green!30}67.9  & \cellcolor{green!30}65.6 &\cellcolor{green!30}61.9  &  \cellcolor{green!30}65.0 \\	\hline 
            \cellcolor{gray!40}Ours  &\cellcolor{gray!40}65.9 &\cellcolor{gray!40}72.2  & \cellcolor{gray!40}65.2& \cellcolor{gray!40}73.0 &\cellcolor{gray!40}68.8  & \cellcolor{gray!40}66.9 & \cellcolor{gray!40}67.7  \\
            \hline
		\end{tabular}
		}
	\end{center}
\end{table}

\noindent\textbf{Results on RGBT234}.~The results on RGBT234 dataset are shown in Table \ref{table:comparison-with-RGBT}.~From the table, we can see that the proposed RGB-T tracker outperforms four supervised trackers and three non-learning-based RGB-T trackers in terms of MSR.~However, our RGB-T tracker shows worse performance on RGBT234 than GTOT.~This is because RGBT234 is much more challenging than GTOT by having more images (223.8K frames v.s.~15.8K frames) and challenging scenarios (12 challenging attributes v.s.~7 attributes).~Most deep RGB-T trackers achieve good performance on RGBT234 by using complex model architectures and a large number of annotated RGB-T image pairs for training.~In contrast, our tracker is trained using 6,900 unlabeled RGB-T image pairs.~In our future work, we will aim to narrow this gap by using better backbone trackers or larger training sets.~However, it is worth mentioning that the performance gap between our self-supervised RGB-T tracker and the state-of-the-art supervised RGB-T tracker (HMFT), i.e., 22.6$\%$ in MPR and 15.2$\%$ in MSR, is acceptable.~This level of gaps also exist between state-of-the-art unsupervised RGB trackers and 
supervised RGB trackers.~For example, the ULAST \cite{shen2022unsupervised} 
achieves 59.2$\%$ in precision and 65.4$\%$ in success rate on the TrackingNet, while SwinTrack \cite{lin2022swintrack} achieves 82.8$\%$ and 84.0$\%$, respectively.

\subsection{Qualitative results}
\label{subsec:qualitative}
\subsubsection{Compared with SOTA trackers}
\begin{figure*}
	\centering
	\includegraphics[width=\textwidth]{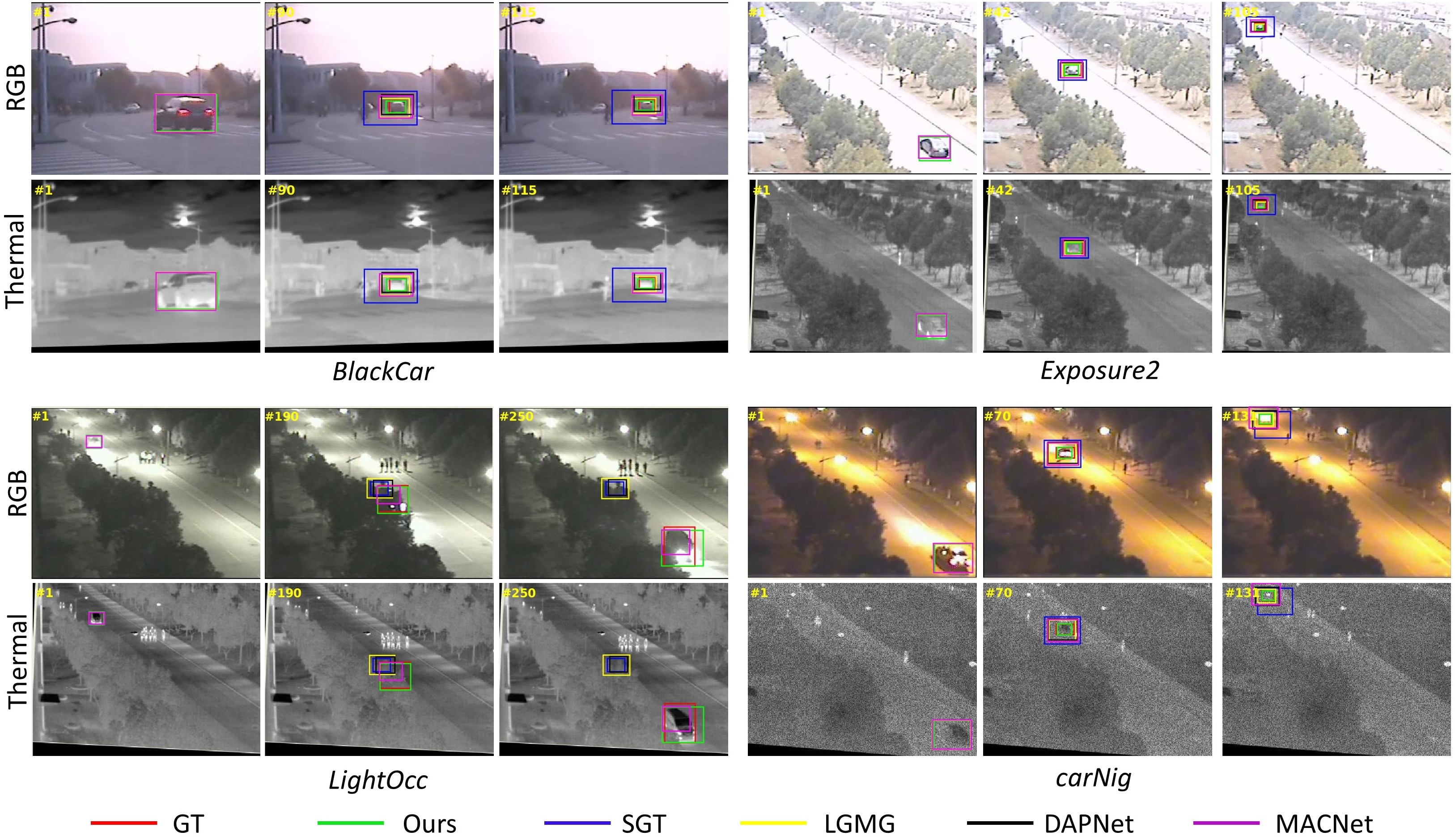}
	\caption{Qualitative comparison of our RGB-T tracker with other RGB-T trackers.~The four sequences are \textit{BlackCar}, \textit{Exposure2}, \textit{LightOcc}, and \textit{carNig} from the GTOT dataset.~We visualize results on both RGB (1st and 3rd rows) and thermal images (2nd and 4th rows).}
		\label{fig:qualitive}
\end{figure*}

We first show qualitative 
comparison of our RGB-T tracker with other RGB-T trackers in Fig.~\ref{fig:qualitive}.~Several RGB-T trackers are selected, including SGT \cite{li2017weighted}, LGMG \cite{li2019learning},  DAPNet \cite{zhu2019dense},  MaCNet \cite{zhang2020object}.~As can be seen, our RGB-T tracker obtains better performance on these four 
sequences, namely, \textit{BlackCar}, \textit{Exposure2}, \textit{LightOcc}, and \textit{carNig}.

\subsubsection{Challenging cases}
\begin{figure}
	\centering
	\includegraphics[width=0.48\textwidth]{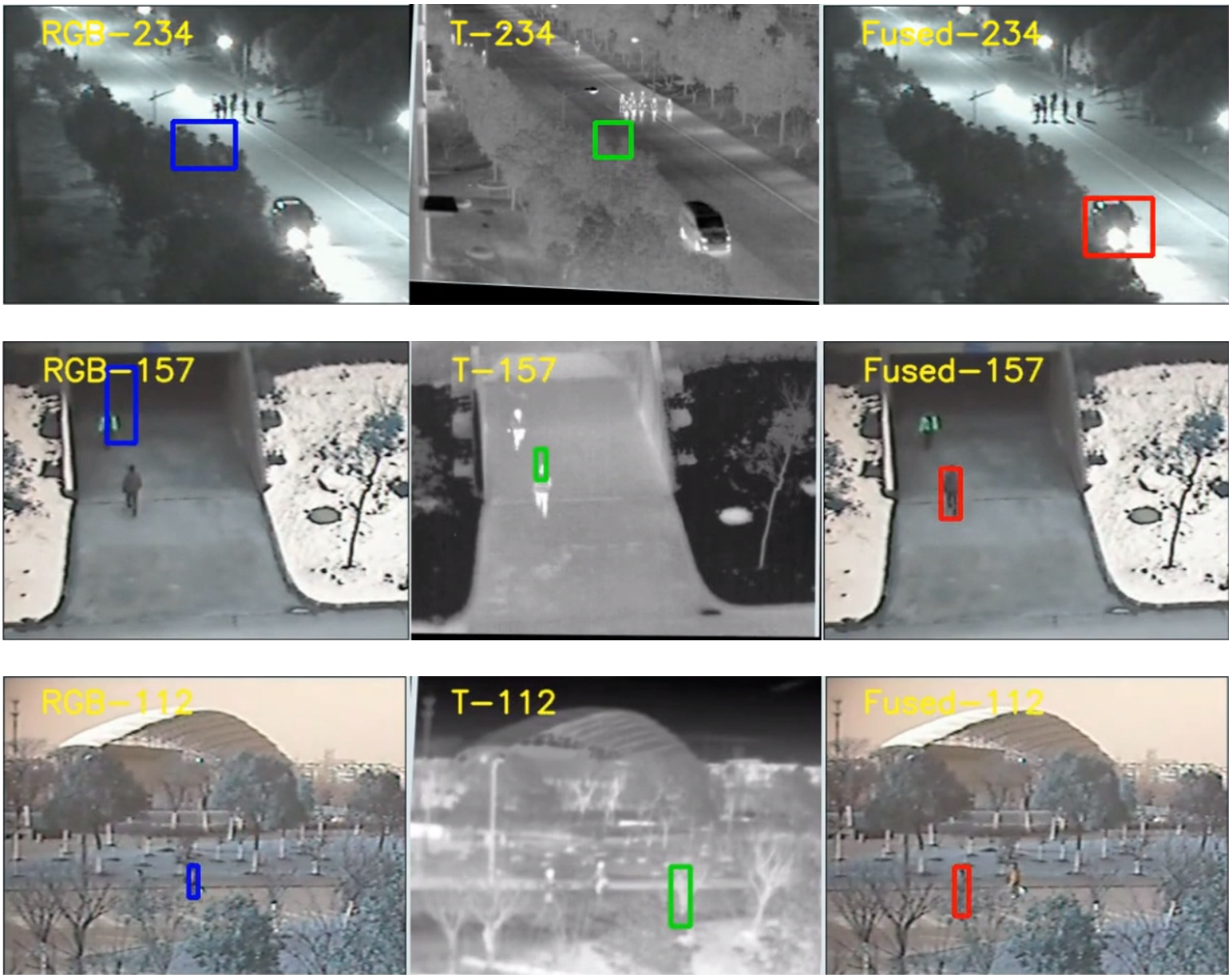}
	\caption{Tracking using RGB images and tracking using thermal images fail.~Our RGB-T tracker can track targets successfully.~From top to bottom: \textit{LightOcc}, \textit{Quarreling}, \textit{WalkingOcc}.~\xzb{Left}: RGB tracker.~\xzg{Middle}: thermal tracker.~\xz{Right}: our self-supervised RGB-T tracker.}
	\label{fig:rgbt-better}
\end{figure}

In this section, we show qualitative results of some challenging cases in Fig.~\ref{fig:rgbt-better}.~In these cases, tracking using RGB images and tracking using thermal images fail, while our RGB-T tracker can correctly track targets.~These examples demonstrate that our self-supervised RGB-T tracker can well fuse RGB and thermal information to improve tracking performance.~For example, in the \textit{LightOcc} case, both the RGB tracker and thermal tracker has problems when the car is heavily occluded by the trees.~In contrast, our self-supervised RGB-T tracker can successfully track the car by fusing RGB and thermal features.~Note that the RGB tracker uses our RGB CNN, and the thermal tracker uses our thermal CNN.

\subsection{Failure cases}
\label{subsec:failure}
\begin{figure}
	\centering
	\includegraphics[width=0.48\textwidth]{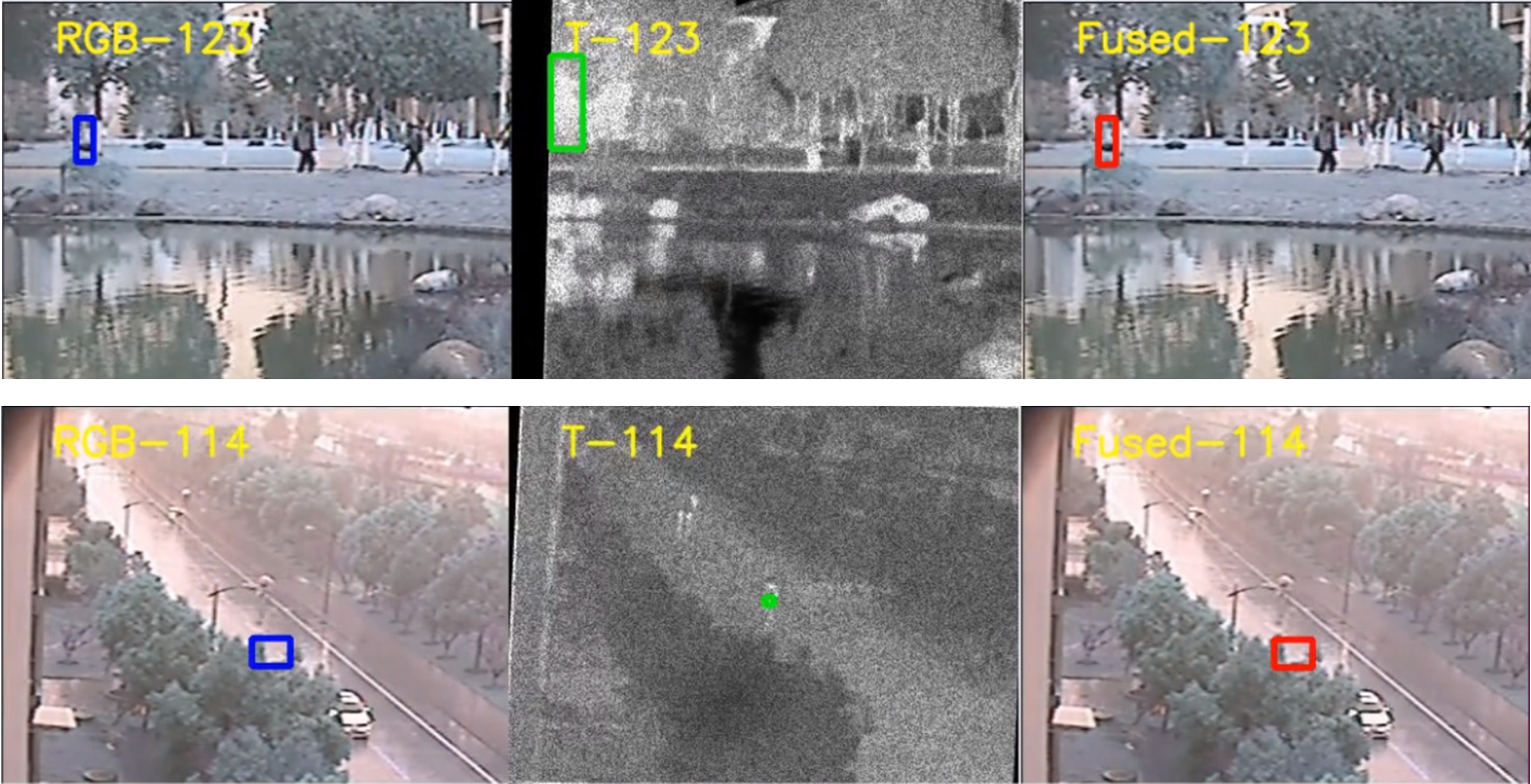}
	\caption{Failure cases of our RGB-T tracker.~Results on \textit{Pool} (top) and \textit{RainyCar2} (bottom) are shown.~\xzb{Left}: RGB tracker.~\xzg{Middle}: thermal tracker.~\xz{Right}: our self-supervised RGB-T tracker.}
	\label{fig:fail}
\end{figure}

Our self-supervised RGB-T tracker fails in some very challenging cases.~Fig.~\ref{fig:fail} shows failure cases of our method.~In these cases, all the RGB tracker, thermal tracker, and RGB-T tracker fail.~In the \textit{Pool} case, the occlusion of the tree and the similar color between tree and pedestrian's cloth are the main reasons of failure.~In the \textit{RainyCar2} case, the occlusion of the trees and the similar color between the white car and road are the main reasons of failure.

\section{Discussion}
\label{sec:discussion}
\textbf{Benefits of our self-supervised training strategy}.~First, the proposed training strategy allows us to train an RGB-T tracker without any human annotations.~Some self-supervised tracking methods \cite{yuan2020self, yuan2021self, li2021self} still require some annotations (very sparse).~Some self-supervised tracking methods use unsupervised optical flow models \cite{zheng2021learning,shen2022unsupervised} to generate pseudo labels or use EdgeBox \cite{wu2021progressive} to generate object proposals.~Our methods totally remove the need for any annotations and optical flow models while achieving reasonable tracking performance, which is very beneficial.~Second, our cross-input consistency is flexible.~Specifically, we can use different modalities as different inputs, which is very suitable for visual data of different modalities, such as RGB images and thermal images.~The idea has the potential to be used in other modalities, e.g., RGB-D tracking.~We can also easily add or remove branches (inputs).~Third, the proposed self-supervised training strategy is generic.~In theory, different backbone trackers can be used in the framework.~We will explore this in the future.

\textbf{Model collapsing}.~A collapsed solution may exist in our cross-input consistency-based method, that is, different branches always generate the same wrong results.~However, this collapsed solution never appeared in our experiments.~Our tracker avoids this issue safely.~A possible reason is that our inputs come from different modalities, so it is unlikely that different branches generate the same features.

\textbf{Limitations}: There is a clear performance gap between our method and state-of-the-art supervised RGB-T trackers.~More efforts should be made to narrow this gap in the future.~A possible solution is to apply our self-supervised training strategy to better backbone trackers, such as SiamAtt \cite{yang2020siamatt}, SiamDW \cite{SiamDW_2019_CVPR}, and ATOM \cite{danelljan2019atom}.~Another promising way is to use larger RGB-T tracking datasets for training.~In addition, we do not learn an IoU net for bounding box regression.~This is a common issue in unsupervised tracking \cite{wang2019unsupervised, wang2021unsupervised, yuan2021self, zhu2021contrastive}.~We will also solve this in our future study.~However, it is worth mentioning that although a performance gap exists between the proposed self-supervised method and state-of-the-art supervised methods, the proposed method has a very good potential because it does not require any manual annotations.

\section{Conclusions}
\label{sec:conclusion}
In this paper, we propose a self-supervised training strategy based on cross-input consistency to train RGB-T trackers.~We construct two distinct inputs using RGB images and thermal images.~Then, we compute the cross-input consistency loss based on the tracking results obtained using these two inputs.~We also propose a loss re-weighting scheme to improve training.~The main benefit of the proposed method is that only unlabeled RGB-T video pairs are needed for training.~Our experiments show that the proposed method achieves favorable performance.~Specifically, with very simple CNNs and a simple feature fusion method, our RGB-T tracker outperforms several supervised RGB-T trackers on the GTOT dataset.~We also show that  the tracking performance can be further improved by using more unlabeled RGB-T videos.~In the future, we will use a more complex network and add a bounding box regression component to the framework.

\section*{Acknowledgments}
This study has received funding from the European Union’s Horizon 2020 research and innovation programme under the Marie Skłodowska-Curie grant agreement No. 101025274.~This work is also funded by a Royal Academy of Engineering Chair in Emerging Technologies to YD.

\small{
\bibliographystyle{IEEEtran}
\bibliography{xingchen}
}
	
 \begin{IEEEbiography}[{\includegraphics[width=1in,height=1.25in]{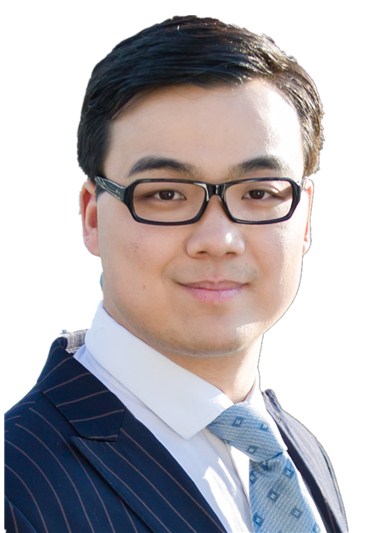}}]{Xingchen Zhang (M'21)} received the B.Sc. degree from the Huazhong University of Science and Technology in 2012, and the Ph.D. degree from the Queen Mary University of London in 2018. He is currently a Marie Skłodowska-Curie Individual Fellow at the Personal Robotics Laboratory, Department of Electrical and Electronic Engineering, Imperial College London. Prior to this, he was a Teaching Fellow and Research Associate at the same department. His main research interests include human intention prediction, image fusion, and object tracking. He is a recipient of the Best Paper Honourable Mention Award of the 9th Chinese Conference on Information fusion.~He is a co-author of the book Image Fusion that has been awarded the National Science and Technology Academic Publications Fund of China (2019).~He is a reviewer for UKRI Future Leaders Fellowship,  EPSRC New Investigator Award and EPSRC Open Fellowship.
	
\end{IEEEbiography}

\begin{IEEEbiography}[{\includegraphics[width=1in,height=1.25in]{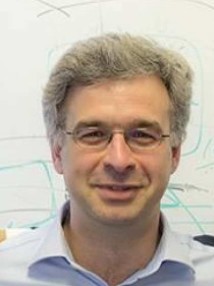}}]{Yiannis Demiris (SM'03)} received the B.Sc. (Hons.) degree in artificial intelligence and computer science 	and the Ph.D. degree in intelligent robotics from the Department of Artificial Intelligence, University of Edinburgh, Edinburgh, U.K., in 1994 and 1999,respectively.~He is a Professor with the Department of Electrical and Electronic Engineering, Imperial College London, London, U.K., where he is the Royal Academy of 	Engineering Chair in Emerging Technologies, and the Head of the Personal Robotics Laboratory. His current research interests include human-robot interaction, machine learning, user modeling, and assistive robotics.~Prof. Demiris was a recipient of the Rector’s Award for Teaching Excellence in 2012 and the FoE Award for Excellence in Engineering Education in 2012. He is a Fellow of the Institution of Engineering and Technology (IET), and the British Computer Society (BCS).
	
\end{IEEEbiography}
	
\end{document}